\documentclass[runningheads]{llncs}

\usepackage{times}
\usepackage{graphicx} 
\usepackage{epsfig} 
\usepackage{amsmath} 
\usepackage{amssymb} 
\usepackage{tabularx}
\usepackage{tabu}
\usepackage{hhline}
\usepackage{stfloats}
\usepackage{array,colortbl}
\usepackage{makecell}
\usepackage{subfigure}
\usepackage{cite} 

\newcommand{\etal}{\textit{et al}. }

\DeclareMathOperator*{\argmin}{arg\,min}

\input{psfig.sty}

\begin{document}

\pagestyle{headings}
\mainmatter

\title{Efficient Hand Articulations Tracking using \\Adaptive Hand Model and Depth map}

\titlerunning{Lecture Notes in Computer Science}

\author{Byeongkeun Kang, Yeejin Lee, and Truong Q. Nguyen}
\institute{Department of Electrical and Computer Engineering, University of California - San Diego}

\maketitle

\begin{abstract}
Real-time hand articulations tracking is important for many applications such as interacting with virtual / augmented reality devices. However, most of existing algorithms highly rely on expensive and high power-consuming GPUs to achieve real-time processing. Consequently, these systems are inappropriate for mobile and wearable devices. In this paper, we propose an efficient hand tracking system which does not require high performance GPUs. 

\quad In our system, we track hand articulations by minimizing discrepancy between depth map from sensor and computer-generated hand model. We also re-initialize hand pose at each frame using finger detection and classification. Our contributions are: (a) propose adaptive hand model to consider different hand shapes of users without generating personalized hand model; (b) improve the highly efficient re-initialization for robust tracking and automatic initialization; (c) propose hierarchical random sampling of pixels from each depth map to improve tracking accuracy while limiting required computations. To the best of our knowledge, it is the first system that achieves both automatic hand model adjustment and real-time tracking without using GPUs.
\end{abstract}

\section{Introduction}
Hands are used in daily lives to handle objects and to better communicate with others. Especially, hands are almost the only way to control electronic devices except limited usage of speech. It is limited since speech is hard to protect privacy, and understanding of speech is difficult in noisy environment. Recent advancements in mobile devices and wearable devices demand better communication methods rather than touch screens which limit physical space. Due to the demand of more natural and convenient interacting methods, interaction using hand gestures has received lots of attention for human-computer interactions, virtual / augmented reality, and robot controls.

\subsection{Related Work} 
Previously, hand pose estimation methods are classified into single frame-based methods and model-based tracking methods~\cite{erol}. Single frame-based methods estimate hand pose by searching huge databases or by recovering hand pose from hand joint classification. Athitsos \etal and Wang \etal used a color image to retrieve hand pose from large databases~\cite{athi, wang}. The method in~\cite{wang} used a color glove for better searching from a database. However, since the database has limited number of hand pose images, it can only estimate the poses in the database. Recently, Tang \etal and Tompson \etal estimated hand pose by applying hand joint classification using random forest and convolutional neural networks (CNNs) respectively~\cite{tang, tompson}. Tang \etal proposed the semi-supervised transductive regression (STR) forest method with joint refinement procedure~\cite{tang}. Tompson \etal employed CNNs and pose recovery to achieve continuous pose estimation~\cite{tompson}. These methods require high performance GPUs to achieve real-time processing and also require large real and synthetic database for training.

Model-based tracking methods estimate hand pose by finding optimal parameters of computer-generated hand model using both current input image and previous results~\cite{sten}. Rehg \etal and Oikonomidis \etal used multiple RGB cameras to reduce occlusions and to increase visual features~\cite{rehg, oikoaccv}. The method in~\cite{oiko} tracks a full DOF hand motion by minimizing discrepancy between input RGB-D image and computer-generated model using particle swarm optimization (PSO). Generating many possible hand pose images for each frame using computer graphics is computationally expensive and requires high performance GPUs to achieve about 15$\sim$20 frames per second (fps) performance. Also, this method requires a user to place a hand on pre-determined position and pose to initialize tracking.

Recently, Sridhar \etal proposed the combined method of single frame-based method and model-based tracking method to blend advantages of each method. They used multiple color cameras for model-based tracking and a depth sensor to search their database, then a voting algorithm is applied to combine the results. Although multiple camera system helps to achieve better accuracy, it requires setup and calibration processes. Moreover, it requires GPUs to process multiple inputs at each frame. Qian \etal proposed another combined method using a depth sensor~\cite{qian}. They combined an efficient initialization method and a tracking method using PSO and iterative closest point (ICP). Their hand model is designed using only spheres to simplify objective function. Sharp \etal also proposed the combined method of two-layer re-initialization using random forest and model fitting using PSO and genetic algorithm~\cite{sharp}.

Most of existing hand tracking systems are rely on expensive, high power-consuming, and high performance GPUs since hand tracking is challenging because of complex articulations, self-occlusions, deformation, and rapid motions. Consequently, these systems are inappropriate for portable and wearable devices. Hand tracking systems for those devices do not need to consider huge viewpoint changes since in general, a user's hand is relatively close to camera. In this paper, we focus on efficient hand articulations tracking system that does not require GPUs and considers mainly the situation when a user's hand is close to camera. Even though we mainly consider small viewpoint changes, it is very challenging because of very limited computational power. Although we design and test our system without using GPUs, it can be implemented with inexpensive and low power-consuming GPUs for better accuracy while still being able to be used for mobile devices. To the best of our knowledge, the proposed system is the first one that can automatically adjust hand shapes which we call adaptive hand model while other methods manually and experimentally decided hand model size for each user. Moreover, we focus on real-time system without using GPUs for mobile and wearable devices. We also propose hierarchical random sampling of pixels on each depth map to achieve better performance with limited computations. Lastly, we improve an efficient re-initialization method at each frame using finger detection and classification.

\begin{figure}
\begin{center}
 \includegraphics[width=0.9\linewidth]{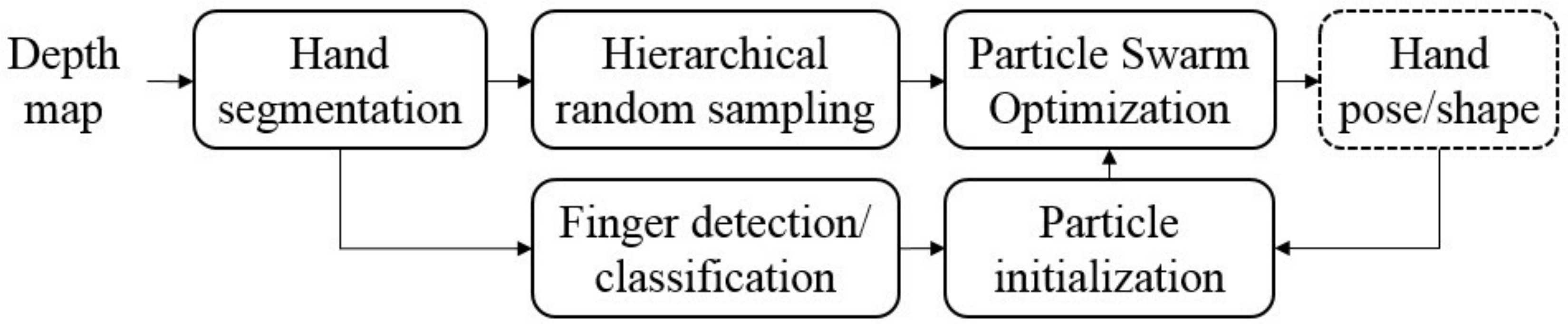}
\end{center}
   \caption{Block diagram of the proposed method.}
\label{fig:block}
\end{figure}

Fig.~\ref{fig:block} shows the entire process of proposed method. We first process hand segmentation from depth map and choose partial pixels from hand region using hierarchical random sampling. We also extract fingertip positions and finger joint rotations using an efficient finger detection and classification algorithm for re-initialization at each frame. Then hand pose and shape are estimated by minimizing discrepancy between the selected partial pixels and computer-generated hand model using PSO. Particles are initialized by previous frame's result and finger detection / classification result.
\vspace{-0.3cm}

\section{Method}
\subsection{Hand Segmentation}
We process a very simple and effective segmentation by using a black wrist band and by assuming that a user's hand is the closest object from camera. This assumption is valid in general environments where one interacts with mobile and wearable devices. The black wrist band is to get depth voids around the wrist since depth sensor cannot capture depth from black object well. The segmentation is processed by finding the connected components from the closest point. For details, we refer the reader to~\cite{kang}.

\subsection{Hierarchical Random Sampling}
We propose hierarchical random sampling of pixels on each depth map for efficient comparison between depth map from sensor and computer-generated hand model. It is computationally expensive to draw computer-generated hand model on image plane and compare entire image to input depth map. To reduce required computations, this process is replaced by comparing subset of pixels on input image to computer-generated hand model with only spheres. Thus we do not need to draw computer-generated hand model on image plane since the difference can be computed without drawing the model. Although tracking accuracy is improved with more subset points, required computation for comparison is also increased. Therefore, we focus on improving the selection of pixels to process from each depth map by applying hierarchical random sampling. The sampling aims to include more pixels on the region which has large depth variations since the region can be interpreted as more informative on depth map.

First, initial samples $S_1$ are randomly sampled. Then, hierarchical sampling $S_2$ is employed to include more points on large depth variation regions. To find large depth variation regions, the gradient matrix $G$ is computed as the sum of absolute x- and y- directional gradient of depth map $D$. The gradients of x- and y- direction are computed by $3 \times 3$ Sobel operators ($O_x$ and $O_y$) in hand segment region:
\begin{equation} 
G = | O_x \ast D | + | O_y \ast D |, 
\end{equation}
where $|\cdot|$ indicates component-wise absolute value and $\ast$ represents convolution. 

For initial samples with large gradient, random samples $S_R$ are selected around initial samples by adding random values $u_1, u_2$ from discrete uniform distribution to x- and y- coordinates respectively: 
\begin{equation}
S_R = \{ S_1 + (u_1, u_2) \mid G(S_1) > t_1 \}. 
\end{equation}
The random samples in $S_R$ are included in hierarchical sample set $S_2$ if the depth difference between initial sample and random sample is greater than a threshold $t_2$:
\begin{equation}
S_2 = \{ S_R \mid | D(S_R) - D(S_1) | > t_2 \}. 
\end{equation}
The final sample set $S$ includes both initial samples $S_1$ and hierarchical samples $S_2$.
Sampled points $S \in \mathbb{Z}^{2\times N_s}$ are converted to $X \in \mathbb{R}^{3\times N_s}$ with $(x,y,z)$ in millimeter, where $N_s$ is the total number of samples.
\begin{figure}
\centering
\subfigure[]{\includegraphics[width=0.2\linewidth]{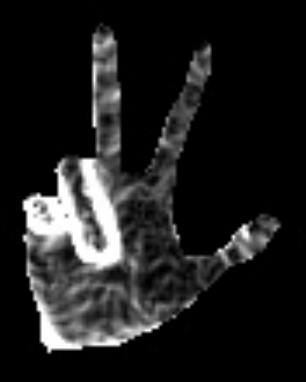}}
~ \quad
\subfigure[]{\includegraphics[width=0.2\linewidth]{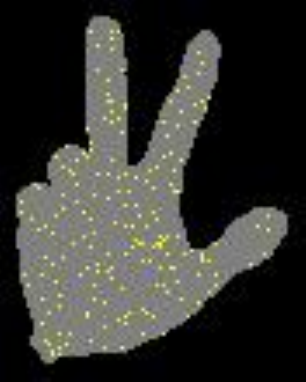}}
~ \quad
\subfigure[]{\includegraphics[width=0.2\linewidth]{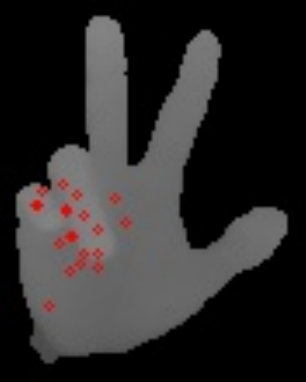}}
~ \quad
\subfigure[]{\includegraphics[width=0.2\linewidth]{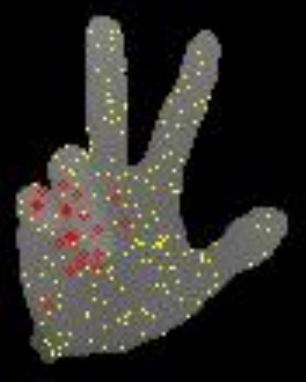}}
\caption{Hierarchical random sampling of pixels from each depth map. This sampling aims to include more pixels on the region which has large depth variations. (a) gradient matrix $G$ of depth map; (b) random sampling $S_1$; (c) hierarchical sampling $S_2$; (d) hierarchical random sampling $S$. }
\label{fig:sampling}
\end{figure}

\subsection{Adaptive Hand Model}
Tracking accuracy increases as the hand model becomes more similar to each user's hand. Although a personalized model can be generated by scanning the user's hand, it requires a pre-processing step for each user. Therefore, we propose the adaptive hand model to consider different hand size and shape while avoiding to scan each user's hand.

\begin{figure}
\centering
\subfigure[]{\includegraphics[width=0.24\linewidth]{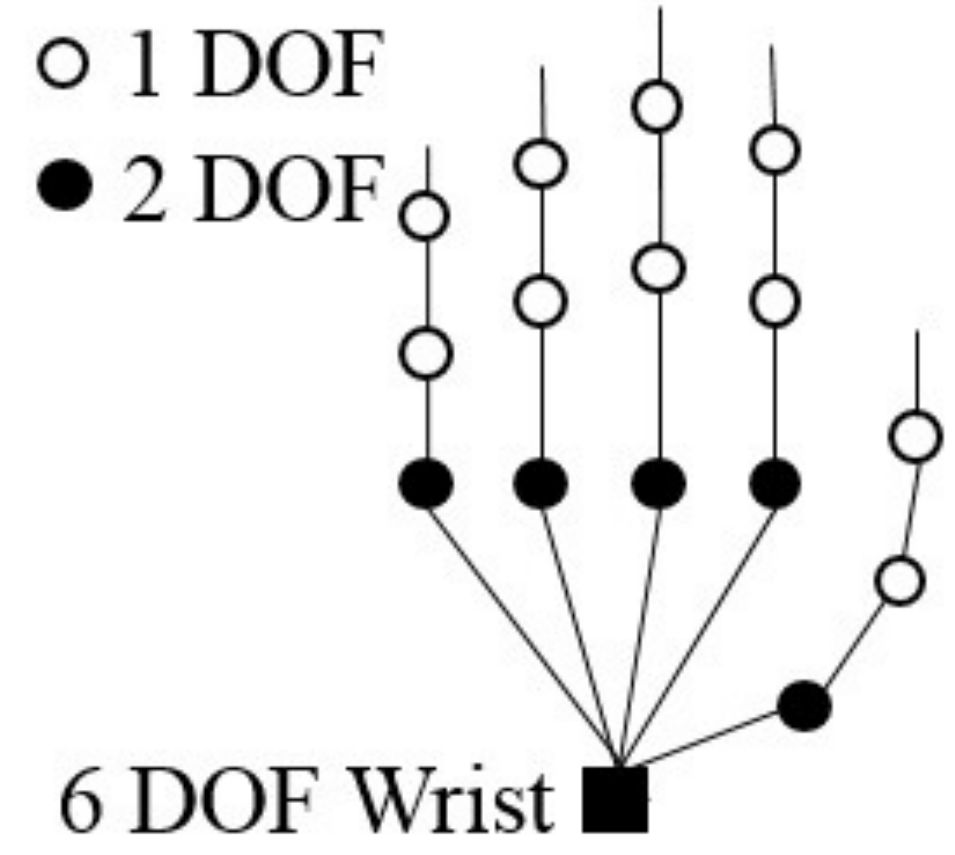}}
~ \qquad
\subfigure[]{\includegraphics[width=0.2\linewidth]{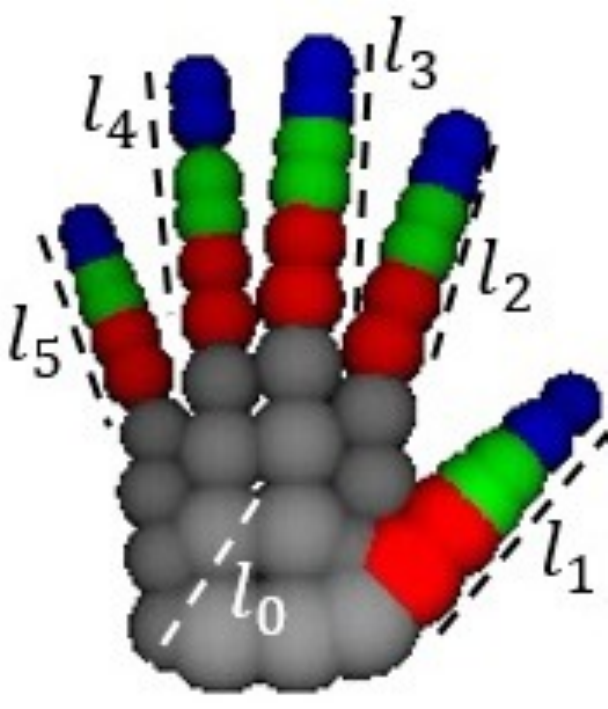}}
   \caption{Hand model. (a) skeletal structure of hand model with 26 DOFs. Each DOF represents a parameter in hand pose parameter vector; (b) an example of hand model with 48 spheres. Two spheres are hidden at connection between palm and thumb. Hand size parameters are visualized.}
\label{fig:hand model}
\end{figure}

Our adaptive hand model consists of a hand size parameter vector $\mathbf{l}\in \mathbb{R}^{6}$ for palm size and finger lengths, and a hand pose parameter vector $\mathbf{p}\in \mathbb{R}^{26}$ for hand position and joint rotations. For hand size parameters, one parameter $l_0$ is for palm size and five parameters $\{l_1, ..., l_5\}$ are for finger lengths from thumb finger to pinky finger. Finger width parameter is not considered since finger width is relatively less important and also computation power is limited. Hand pose parameter vector is defined to estimate translations and rotations of hand joints as in Fig.~\ref{fig:hand model} (a).

Since hand shape at each frame is dependent on both size and pose parameters, two parameter vectors should be optimized simultaneously. However, the combination of two parameter vectors is 32 parameters, which is really complex to optimize even without considering correlation between parameters. Therefore, size parameters are considered only when five fingers are detected and classified by the re-initialization method in Section 2.5 since in that case, the optimization of pose parameters is relatively accurate and robust. 

Hand model is designed using only spheres to reduce the computational complexity of objective function~\cite{qian}. The main reason is that in comparison between input depth map and the hand model, the length from a point to the surface of a sphere is simply the length from the point to the center of sphere subtracted by the radius. 

\subsection{Optimization}
\subsubsection{Objective Function}
Our objective function is designed mainly to minimize Euclidean distance between sampled pixels from input depth map and computer-generated hand model. The function consists of two discrepancy terms between hand model and depth map and one validity term. Two discrepancy terms are the Euclidean distance from depth map to hand model and from the model to depth map. The validity term is to check invalid overlapping between parts of hand model. The total cost is the weighted summation of these three terms. For details of general objective function, we refer the reader to the paper~\cite{qian}. For our system, since hand shape at each frame is determined by both hand pose and hand size parameters, cost is also determined by both parameters. We conducted experiment with another objective function which incorporates additional cost term to minimize temporal hand size parameter changes. However, tracking accuracy is not improved since the variation of hand size parameters are already regularized by the normal distribution.

\subsubsection{Particle Swarm Optimization}
Modified Particle Swarm Optimization (PSO) is used to find the best hand pose parameters and hand size parameters by minimizing an objective function. Each particle represents one state of hand pose and hand size parameters in our algorithm. The optimization method first initializes particles over possible solution range, and then finds the best solution among particles using objective function. Then particles are moved from current state to the direction of the best solution. The algorithm iterates finding the best solution and moving particles to the direction of best solution until it reaches maximum generation or termination condition.

Entire particles are initialized as the sum of optimized parameters $(\mathbf{p}_o^{(t-1)}, \mathbf{l}_o^{(t-1)})$ at previous frame and random values $(\mathbf{r}_p, \mathbf{r}_l)$ from normal distribution if corresponding finger is not detected and classified by the method in Section 2.5. Otherwise, 75\% of the corresponding parameters are initialized with the same method, and 25\% of them are initialized with the sum of measured parameters from Section 2.5 and random values. The distribution of random values are $\mathbf{r}_p \sim \mathcal{N} (0, \boldsymbol{\sigma}_1^2)$ and $ \mathbf{r}_l \sim \mathcal{N}(0,\sigma_2^2)$. 
However, at the first generation, hand size parameters are not considered to focus on the optimization of pose parameters since inaccurate pose parameters lead to wrong size parameters and the size parameters from previous frame are relatively reliable.

At each generation, for pose parameters, particles are updated using global best particle $\mathbf{g}$ and personal best particle $\mathbf{b}$. Personal best particle is the state when the particle has the lowest cost until current generation. Global best particle is the lowest cost state among all particles and all generations until current generation. Particles are updated to the direction of personal best particle and global best particle using the following rules:
\begin{equation}
\begin{aligned}
& \mathbf{b}_{i,j} = \{ \mathbf{p}_{i,\tilde{k}} | \tilde{k} = \argmin_{k} {C(X^{(t)}, \mathbf{l}_{i,1}^{(t)}, \mathbf{p}_{i,k}^{(t)} )}  \}, \\
& \mathbf{g}_{j} = \{ \mathbf{p}_{\tilde{i},\tilde{k}} | (\tilde{i},\tilde{k}) = \argmin_{i,k} {C(X^{(t)}, \mathbf{l}_{i,1}^{(t)}, \mathbf{p}_{i,k}^{(t)} )} \}, \\
& \mathbf{p}_{i,j} = \mathbf{p}_{i,j-1} + \alpha_1 (\mathbf{b}_{i,j-1} - \mathbf{p}_{i,j-1})  + \alpha_2 (\mathbf{g}_{j-1} - \mathbf{p}_{i,j-1}), 
\end{aligned}
\end{equation}
where $C(\cdot)$ is the objective function, weight $\alpha_1 \sim U[0.5, 1.5]$, weight $\alpha_2 = 2-\alpha_1$, $i$ is the index of each particle, $j$ and $k$ denote the generation index, and $t$ is current frame index.

For size parameters, particles are not updated to avoid misleading caused by the dependency between size parameters and pose parameters.
Although size parameters are not updated at each generation, the particles with better size parameters are likely to have lower cost after many generations since pose parameters will become similar.

After reaching maximum generation, both size parameters and pose parameters are updated with the global best particle.
\begin{equation}
(\mathbf{p}_o^{(t)}, \mathbf{l}_o^{(t)}) = \{ (\mathbf{p}_{\tilde{i},\tilde{k}}^{(t)}, \mathbf{l}_{\tilde{i},1}^{(t)}) | (\tilde{i},\tilde{k}) = \argmin_{i,k} {C(X^{(t)}, \mathbf{l}_{i,1}^{(t)}, \mathbf{p}_{i,k}^{(t)} )} \}. 
\end{equation}

\subsection{Re-initialization at each frame}
Re-initialization at each frame is important to avoid manual initialization and error accumulation. However, general re-initialization methods using random forest or CNNs require large computation load which needs high performance GPUs to achieve real-time. Therefore, we improved the efficient finger detection and classification method proposed by~\cite{qian}. Although this re-initialization method works in limited cases, it initializes hand pose automatically, improves tracking accuracy, and is incorporated in real-time tracking without GPUs. We improve the re-initialization by estimating palm orientation using both current measurement and prior knowledge from previous frames, which is inspired by Kalman filtering~\cite{kalman}.  

\subsubsection{Finger Detection}
A simple finger detection algorithm is employed to detect planar fingers and orthogonal fingers. We define planar fingers as the fingers which are parallel to image plane and orthogonal fingers which are orthogonal to image plane. First, palm center is measured as the maximum of distance transform of hand segment. Then a planar finger candidate is detected by finding connected component from extreme distance point from palm center until the component reaches finger length. The detected finger candidate is classified into either a finger or a non-finger based on the component size. This process is iterated until it detects five fingers or the segment does not have any extreme distance point. After detecting planar fingers, an orthogonal finger candidate is detected by finding connected component from the closest point from a camera on both depth map and hand segment within a small window. It is also classified to either a finger or a non-finger based on the size of region. This process is also iterated until it reaches same condition as planar finger case. For planar fingers, principal component analysis (PCA) is applied to each detected region in order to calculate the orientation of each detected finger. For orthogonal fingers, the orientation is assumed that it is orthogonal to image plane. 
\begin{figure}
\centering
\subfigure[]{\includegraphics[width=0.2\linewidth]{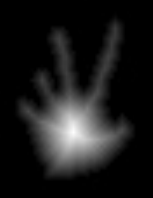}}
~ \qquad
\subfigure[]{\includegraphics[width=0.2\linewidth]{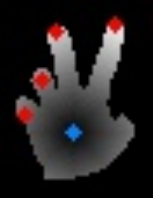}}
~ \qquad
\subfigure[]{\includegraphics[width=0.2\linewidth]{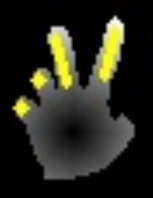}}
~ \qquad
\subfigure[]{\includegraphics[width=0.2\linewidth]{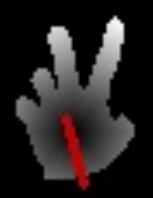}}
~ \qquad
\subfigure[]{\includegraphics[width=0.2\linewidth]{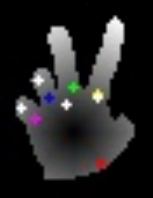}}
~ \qquad
\subfigure[]{\includegraphics[width=0.2\linewidth]{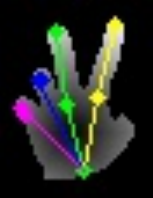}}
\caption{Finger detection and classification. (a) distance transformation; (b) detection of fingertips (red) and palm center (blue); (c) computed finger orientation using PCA; (d) estimated palm orientation; (e) joint junctions (white: detected, colored: predicted using hand model and palm orientation); (f) classification result.}
\label{fig:fingerDetection}
\end{figure}

\subsubsection{Finger Classification}
Finger classification is to use the result from finger detection for particle initialization process in Section 2.4. The algorithm classifies each detected finger to one of five finger classes. 
First, palm orientation is measured by applying PCA to palm segment. However, the measured orientation $\theta_m$ is not robust and accurate enough to use directly for classification. Therefore, palm orientation ${\theta}_{p,t}$ is predicted by the weighted summation of previously estimated palm orientation $\theta_{o,t-1}$ from PSO and currently measured palm orientation $\theta_{m,t}$. The weights in summation are decided by the measurement error $e_{m}$ and priori estimation error $e_{o}$. This is inspired by Kalman filtering~\cite{kalman}. 
\begin{equation}
\begin{aligned}
&{\theta}_{p,t} = \frac{e_{m,t}}{e_{m,t}+e_{o,t}} (\theta_{o,t-1}+c_{t-1}) +  \frac{e_{o,t}}{e_{m,t}+e_{o,t}} \theta_{m,t}, \\
&e_{m,t} = | \theta_{o,t-1} - \theta_{m,t-1} | + a_1 e_{m,t-1}, \\
&e_{o,t} = | \theta_{o,t-1} - (\theta_{o,t-2} + c_{t-2}) | + a_2 e_{o,t-1}, 
\end{aligned}
\end{equation}
where $t$ is current frame index and $a_i$ is chosen as a constant for simplicity. In our experiments, both $a_1$ and $a_2$ are set to 0.5.
After each frame, priori estimation constant $c$ is updated as $c_{t} = \theta_{o,t} - \theta_{o,t-1}$.

A set $F$ of junction points of fingers and palm is computed using detected fingertips, finger orientations, and finger length.
Another set $Q$ of junction points is calculated using hand model and predicted palm orientation.
Each junction point $f_i \in F$ from an input image is matched to the closest junction point $q_j \in Q$ from hand model.
\begin{equation}
J(f_i) = \argmin_{j} ||f_i - q_j||_2. 
\end{equation}
If more than two of detected junctions are classified into the same class, optimization is employed to find the result with minimum cost.
The algorithm first finds possible combinations $B$ that have minimum changes in initial classification.
\begin{equation}
\begin{aligned}
B_k = \{(J(f_1)+v_1,...,J(f_n)+v_n) \mid \min \sum_{i=1}^{n} |v_i|^2, \forall i \neq j, J(f_i) + v_i \neq J(f_j) + v_j \}.
\end{aligned}
\end{equation}
The final class $L$ is classified as follows:
\begin{equation}
L = \{ B_{\tilde{k}} \mid \tilde{k}=\argmin_{k} \sum_{i=1}^{n}(||f_i - q_{B_{k}(i)} ||_2) \},
\end{equation}
where $i$ is the index of detected finger and $n$ is the number of detected fingers.
Fig.~\ref{fig:fingerDetection} (d)-(f) illustrate the procedure of finger classification. In Fig.~\ref{fig:fingerDetection} (e), white circles represent detected finger junctions and colored circles from red to purple indicate hand model junctions from thumb finger to pinky finger. A yellow circle which corresponds to predicted joint junction of index finger is overlapped with a white circle. Although this re-initialization method cannot detect all the fingers at every frame, it improves tracking accuracy. It can also initialize at each frame including the very first frame, and takes only a few milliseconds using only CPU. 

\section{Experimental Results}
The algorithm is tested using a Creative Senz3D camera and a computer with Intel Core i7-3770 3.4GHz CPU, 16GB RAM, and without GPUs. Although the machine has 16GB RAM, this algorithm only uses about 60MB memory. Although a 3.4GHz CPU is used, we believe similar computation power can be obtained using the combination of mobile CPU and mobile GPUs. 

We captured 500 frames for each subject and labeled wrist and five fingertips of last 400 frames. The dataset is available on our repository\footnote{\url{https://github.com/byeongkeun-kang/HandTracking}}. Initial 100 frames contain open-hand pose that camera can capture at least some fingers using the algorithm in Section 2.5 for automatic initialization. The initial frames are not used to compute accuracy. Error in accuracy is computed using 3D Euclidean distance in millimeter. 

Unless specifically mentioned, we sample 256 points from each depth map and optimize with 256 particles and 6 generations.  Table~\ref{tab:processingTime} shows the processing time at this setting. Even though the algorithm is not fully optimized, it achieves about 16 FPS using eight threads on CPU. 
\begin{table}
\begin{center}
\caption{Processing time.}
\label{tab:processingTime}
\begin{tabu} to \textwidth {X[c]||X[c]}
\Xhline{3\arrayrulewidth}
Process & Time (in $ms$) \\
\hline\hline
Finger detection/classification & 2.5 \\
Optimization & 53.5 \\
Others & 6.5 \\
\hline
Total  & 62.5 \\
\Xhline{4\arrayrulewidth}
\end{tabu}
\end{center}
\end{table}

\begin{figure}
\centering
\subfigure[]{\includegraphics[width=0.73\linewidth]{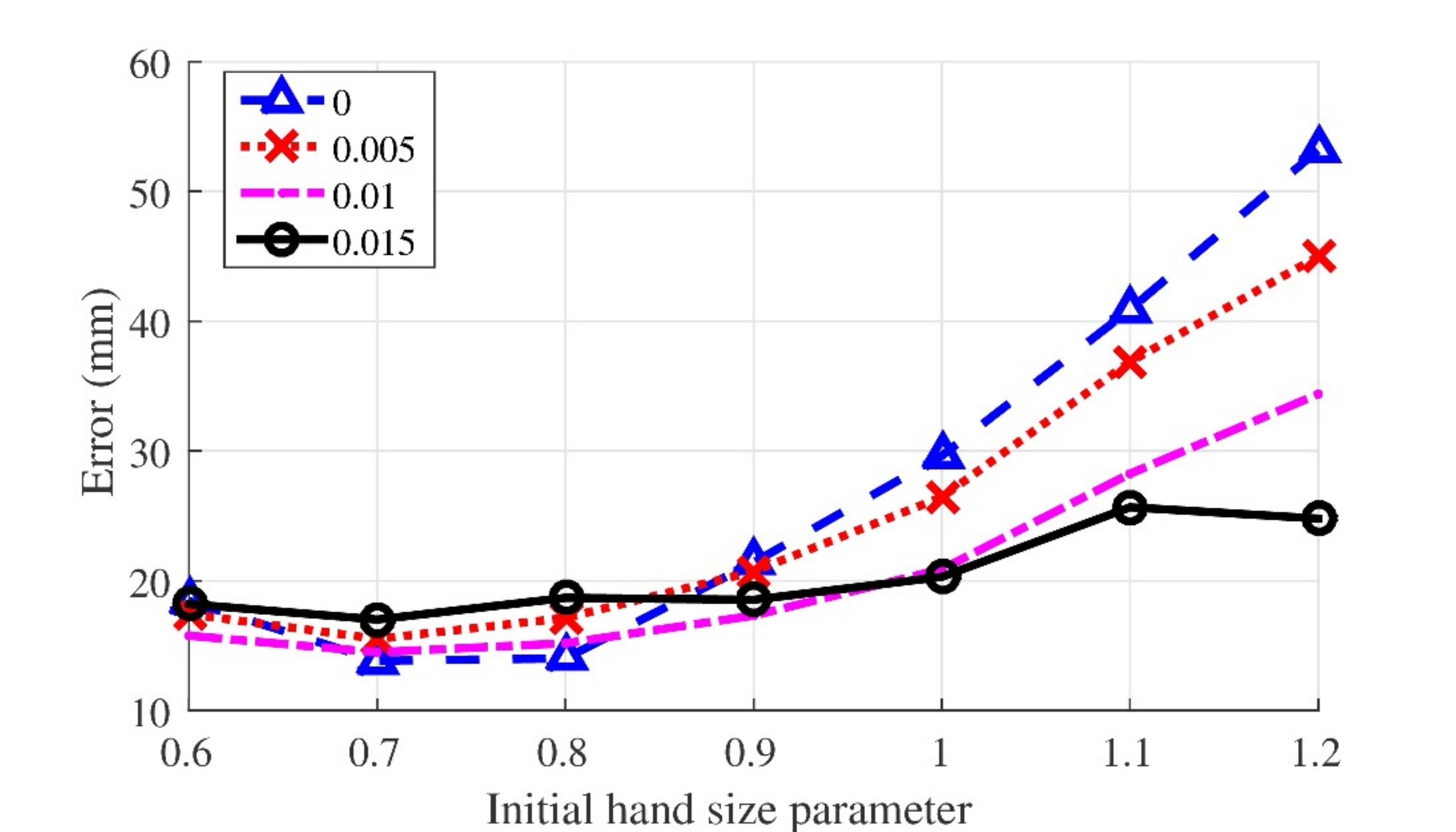}}
\subfigure[]{\includegraphics[width=0.73\linewidth]{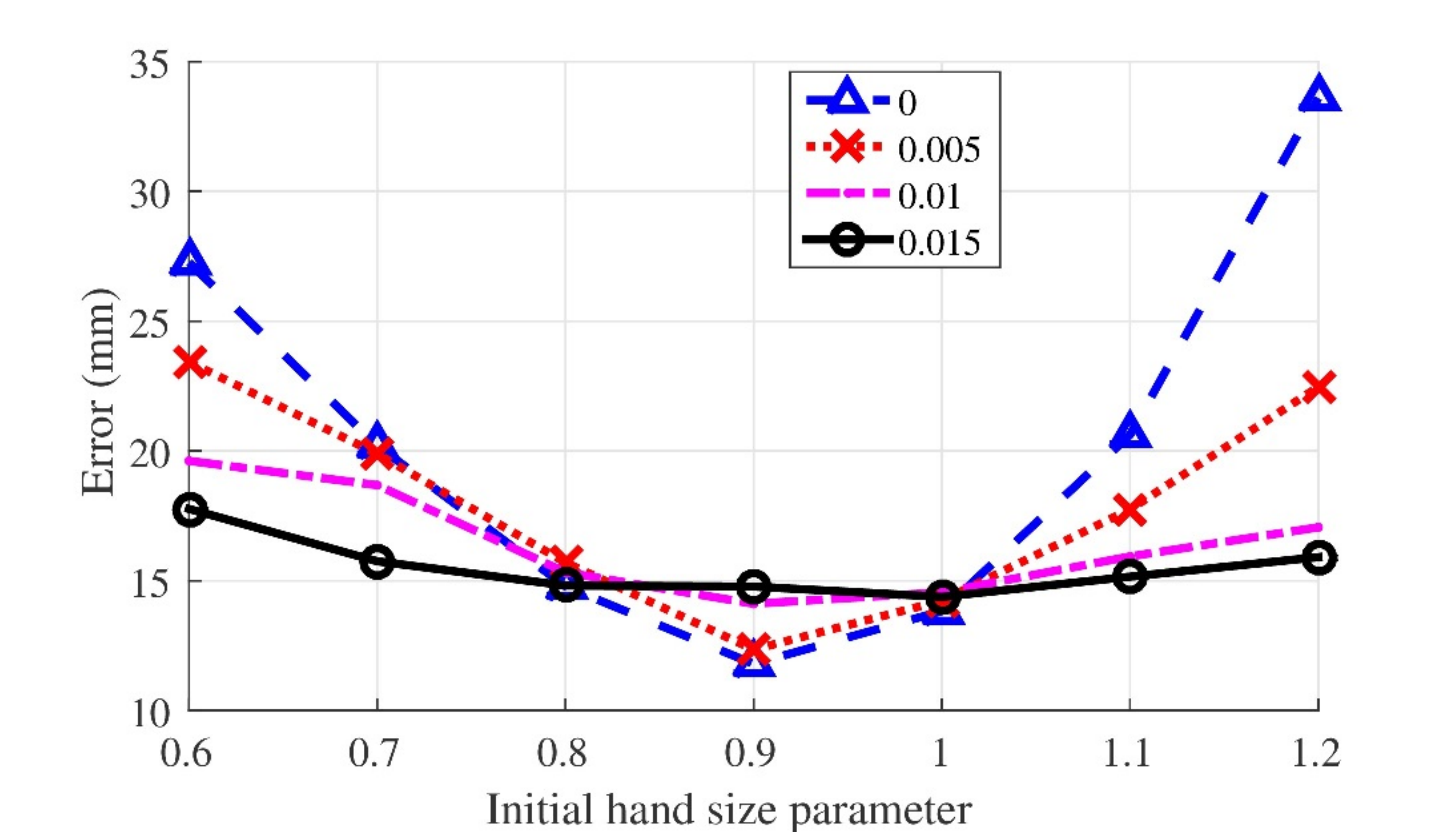}}
\subfigure[]{\includegraphics[width=0.73\linewidth]{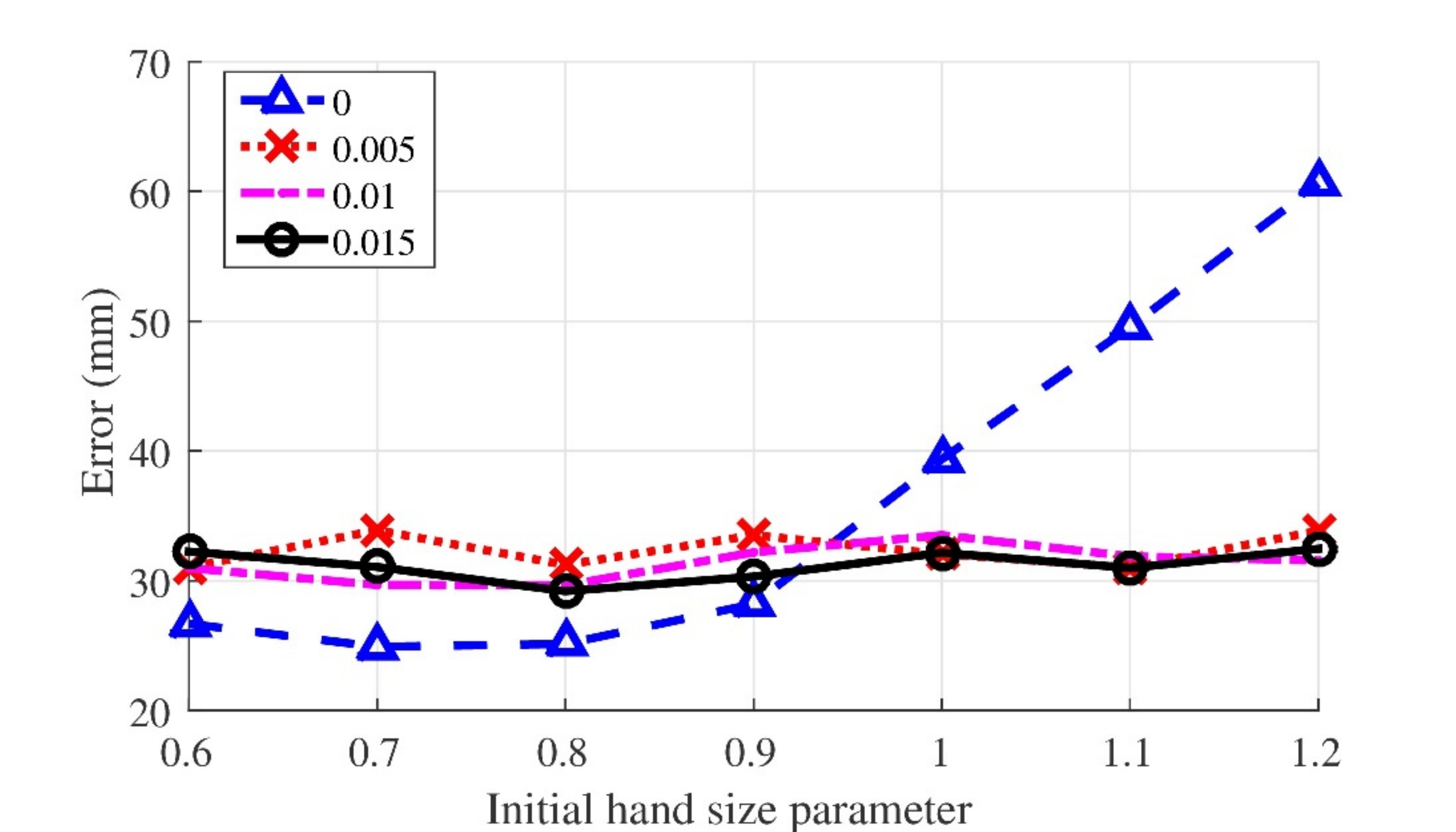}}
\caption{Comparison of adaptive hand model and fixed scale hand model for three subjects (top: subject 1, middle: subject 2, bottom: subject 3). The legend represents standard deviations $\sigma_2$ of hand size parameter randomness. Standard deviation of 0 means that the scaling factor of hand model is always the same with initial scaling factor. Larger standard deviation means more possibility of large update of scaling factor at each frame. The result shows that the performance of adaptive hand model is better than the performance of fixed scaled hand model in general.}
\label{fig:adaptiveResult}
\end{figure} 
\subsection{Adaptive Hand Model}
The performance comparison of hand model with adaptive scaling and fixed scaling is demonstrated in Fig.~\ref{fig:adaptiveResult} for different subjects, initial scaling factors, and hand size parameter randomness. Scaling factors are chosen from 0.6 to 1.2 with 0.1 step size and standard deviation of randomness $\sigma_2$ is chosen from 0 to 0.015 with 0.005 step size. Also, Table~\ref{tab:adaptive} clearly shows that adaptive hand model reduces error about 5$mm$. The overall results show that adaptive hand model adjusts hand scaling factor automatically to minimize discrepancy between pre-defined hand model and user's hand, and users do not need to manually and experimentally select scaling factor of hand model. Moreover, to consider the case that the user is changed after starting tracking, we keep update hand scaling factor.
\begin{table}
\begin{center}
\caption{Average error for different standard deviations $\sigma_2$ of hand size parameter randomness. To compute average error, we consider three subjects and seven initial scaling factors chosen from 0.6 to 1.2. The detailed explanation of standard deviation is on Fig.~\ref{fig:adaptiveResult}.}
\label{tab:adaptive}
\begin{tabu} to \textwidth {X[c]||X[c]|X[c]|X[c]|X[c]}
\Xhline{3\arrayrulewidth}
$\sigma_2$ in Sec. 2.5 & 0  & 0.005 & 0.01 & 0.015 \\
\hline\hline
Error (in $mm$) & 28.00 & 25.31 & 22.90 & 22.38 \\
\Xhline{4\arrayrulewidth}
\end{tabu}
\end{center}
\end{table}

\subsection{Finger Classification}
Correct classification rate (CCR) is calculated for each finger with and without the proposed palm orientation prediction in Table~\ref{tab:CCR}. The result shows that by using the proposed prediction, the average CCR is improved from 70.6\% to 84.8\%. 
\begin{table}
\begin{center}
\caption{Correct classification rate (in $\%$) of finger classification.}
\label{tab:CCR}
\begin{tabu} to \textwidth {c||X[c]|X[c]|X[c]|X[c]|X[c]|X[c]}
\Xhline{3\arrayrulewidth}
CCR & Thumb  & Index & Middle & Ring & Pinky & Average \\
\hline\hline
Without prediction & 87.1 & 65.5 & 69.4 & 59.0 & 71.8 & 70.6 \\
With prediction & 97.2 & 86.3 & 78.1 & 70.0 & 92.1 & 84.8 \\
\Xhline{4\arrayrulewidth}
\end{tabu}
\end{center}
\end{table}

\subsection{Hierarchical Random Sampling}
The performance of random sampling and hierarchical random sampling is compared in Table~\ref{tab:sampling}. The average error is computed using the hand model scaled by the best performance fixed scaling factor, 0.7, 0.9, and 0.7 for subject 1, 2, and 3 respectively. The average accuracy is improved from 17.01$mm$ to 16.11$mm$. The computational cost of calculating gradient is much smaller than increasing the number of generations, particles, or samples to achieve the same improvement. However, if the number of sampling pixels is too small, random sampling might be better since hierarchical random sampling prevents that sampled pixels are more globally distributed.
\begin{table}
\begin{center}
\caption{Performance comparison of random sampling~\cite{qian} and the proposed hierarchical random sampling.}
\label{tab:sampling}
\begin{tabu} to \textwidth {X[c]||X[c]|X[c]}
\Xhline{3\arrayrulewidth}
Sampling & Random~\cite{qian} & Hierarchical Random \\
\hline\hline
Error ($mm$) & 17.01 & 16.11 \\
\Xhline{4\arrayrulewidth}
\end{tabu}
\end{center}
\end{table}

\section{Conclusion}
We present an efficient hand articulations tracking system for mobile and wearable devices which do not have high performance GPUs. We show that the proposed system achieves both automatic hand model adjustment using adaptive hand model and real-time tracking without using GPUs. We also achieve improved accuracy using hierarchical random sampling and improved efficient re-initialization at each frame while limiting required computations.
\bibliographystyle{splncs}
\bibliography{isvc_submission}

\end{document}